\documentclass[letterpaper, 10 pt, journal, twoside]{IEEEtran}
\IEEEoverridecommandlockouts

\usepackage[T1]{fontenc}
\usepackage{times}
\usepackage{amsmath,amssymb,amsfonts}
\usepackage{algorithm}
\usepackage{graphicx}
\usepackage{textcomp}
\usepackage{xcolor}
\usepackage{multirow}
\usepackage{comment}
\usepackage{xfrac}
\usepackage{bm}
\usepackage{microtype}
\usepackage{textcomp}

\DeclareMathOperator*{\argmax}{argmax}

\graphicspath{{figs/}}

\begin{document}
\title{Unsupervised Complementary-aware\\ Multi-process Fusion for Visual Place Recognition}
\author{
Stephen Hausler \qquad \qquad Tobias Fischer \qquad \qquad Michael Milford\\
\thanks{This work has been partially supported by funding from an Amazon Research Award, grant 81418190, and from the Australian Government via grant AUSMURIB000001 associated with ONR MURI grant N00014-19-1-2571. The authors acknowledge continued support from the Queensland University of Technology (QUT) through the Centre for Robotics.}
\thanks{All authors are with the QUT Centre for Robotics, Queensland University of Technology, Brisbane, QLD 4000, Australia.
{\texttt{\{stephen.hausler, tobias.fischer, michael.milford\}@qut.edu.au}}.}
}

\maketitle
\thispagestyle{empty}
\pagestyle{empty}

\begin{abstract}

A recent approach to the Visual Place Recognition (VPR) problem has been to fuse the place recognition estimates of multiple complementary VPR techniques simultaneously. However, selecting the optimal set of techniques to use in a specific deployment environment a-priori is a difficult and unresolved challenge. Further, to the best of our knowledge, no method exists which can select a set of techniques on a frame-by-frame basis in response to image-to-image variations. In this work, we propose an \emph{unsupervised} algorithm that finds the most robust set of VPR techniques to use in the current deployment environment, on a \emph{frame-by-frame} basis. The selection of techniques is determined by an analysis of the similarity scores between the current query image and the collection of database images and does not require ground-truth information. We demonstrate our approach on a wide variety of datasets and VPR techniques and show that the proposed dynamic multi-process fusion (Dyn-MPF) has superior VPR performance compared to a variety of challenging competitive methods, some of which are given an unfair advantage through access to the ground-truth information.

\end{abstract}

\section{Introduction}

Visual Place Recognition (VPR) is an important component for robotic and autonomous systems: it provides the ability to recognise previously visited locations using just image data, which is beneficial for a range of downstream tasks including Simultaneous Localization and Mapping and navigation \cite{Garg2021}. However, VPR is challenging due to the wide variety of appearance and viewpoint changes that can occur in a navigation task. Appearance changes can be as diverse as seasonal changes, such as summer to winter, or illumination changes as challenging as day to night.

Because of the wide range of potential appearance and viewpoint variations, it is challenging to design a ``one size fits all'' approach to VPR that works across all possible deployment environments \cite{Garg2021}. This has been observed in prior literature~\cite{schubert2021makes}, with strong techniques such as NetVLAD \cite{Arandjelovic2018} performing poorly on the Nordland dataset \cite{Sunderhauf2013}, or Histogram of Oriented Gradients \cite{DN2005} failing on the Berlin Kudamm dataset \cite{Hausler2020a}. The recent VPR-Bench framework \cite{Zaffar2021} and other works \cite{schubert2021makes} also showed that the optimal VPR technique for a given dataset varies significantly.

\begin{figure}[t!]
    \centering
    \includegraphics[width=0.99\columnwidth,trim=0cm 4.4cm 10.4cm 0cm,clip]{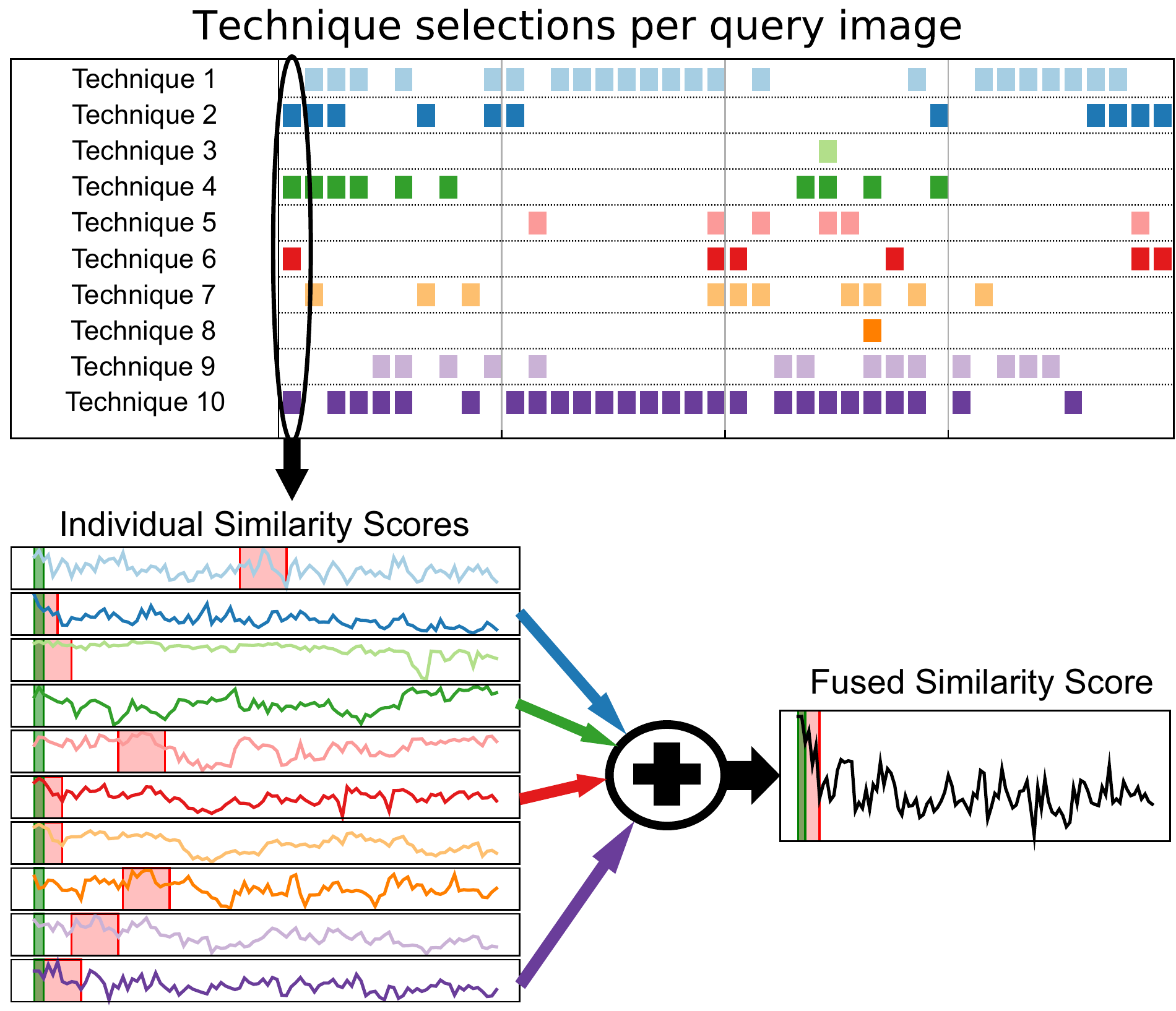} %
    \caption{Given a collection of different VPR techniques, our approach determines which subset of techniques to fuse in order to maximize VPR performance, and updates this subset for each new query image. Our approach is unsupervised and uses an estimate of perceptual aliasing through analysis of the similarity vector between a query and a set of database images. \emph{Top:} we display a demonstration of the techniques selected for a subset of one of the datasets. \emph{Bottom:} for the first query image, we show the similarity scores for each technique and show the fused similarity vector for the chosen techniques. The green bar denotes the ground truth (with the associated ground truth tolerance window) and the red bar denotes the best matching database image (with the associated region of self-similarity).
    }
    \label{fig:demo}
\end{figure}

To resolve this difficulty, a recent idea \cite{SRAL, Hausler2019} has been to fuse multiple VPR techniques in tandem, creating a pseudo-multi-sensor fusion with a single sensor. Fusing multiple sources of information has had significant prior research in multi-sensor fusion applications \cite{Milford2013a, Zhang}. By using multiple VPR techniques simultaneously, limitations with one technique (in a specific environment) can be offset with the inclusion of additional techniques that may not have such limitations. More recent work has shown that certain combinations of techniques are more effective than others on specific datasets -- a phenomenon that has been termed the \emph{complementarity} of different combinations of techniques \cite{Hausler2020a, Waheed2021}.

In this work, we propose an \emph{unsupervised} technique to automatically estimate the optimal set of techniques to fuse in a multi-process fusion algorithm, which \emph{continuously updates as the environment changes} over a deployment traverse. Our algorithm computes an estimate of the perceptual aliasing of different sets of fused techniques, and selects the set of techniques that minimises the perceptual aliasing. We estimate the perceptual aliasing using the ratio between the largest similarity score and the second largest similarity score, within the set of scores between the current query image and the database images \cite{MM2012}. We demonstrate our approach on the 12 VPR-Bench datasets and outperform full multi-process fusion and hierarchical multi-process fusion, both of which using static sets of techniques. Our proposed solution enables multi-process fusion to be deployed in any environment, without requiring calibration or manual selection of techniques. Additionally, no ground-truth information or additional training is required; the selection process is determined purely using the distribution of the matching scores of the different techniques.

We claim the following contributions:
\begin{enumerate}
    \item We propose an \emph{unsupervised} algorithm for automatically selecting a complementary set of VPR techniques to use in a multi-process fusion VPR system, which is updated on an image-by-image basis as the autonomous platform traverses an environment (see Figure \ref{fig:demo}).
    \item We leverage an existing technique \cite{MM2012, Hausler2019} for estimating the precision of a place recognition match, which calculates the similarity score ratio between the most similar database image and the second most similar image. We use this ratio as an estimate of the perceptual aliasing and find the set of techniques that maximizes this ratio.
    \item Our experimental results and ablations demonstrate that the proposed algorithm is able to adjust the selection of techniques on an image-by-image basis, and, on average, our proposed technique outperforms a range of VPR baselines including multi-process fusion, hierarchical multi-process fusion, and the best performing individual technique as determined by an oracle.
\end{enumerate}    

\section{Related Work}

\subsection{Fusing Multiple Modalities for Improved VPR}
Fusing multiple sources of information is a common technique in both SLAM and VPR~\cite{cadena2016past,Garg2021}. Traditionally this has been achieved through the use of multiple physical sensors, like LiDAR, depth cameras, RADAR and Wi-Fi \cite{JA2018}. Numerous algorithms have been devised to intelligently fuse these different sensors and produce a more robust localization estimate \cite{Zhang, Milford2013a}. 

Recent work has shown that some of the benefits of multi-sensor fusion can still be attained with a single sensor, by fusing different interpretations of the image data that are obtained by applying different image processing techniques on the same image. SRAL \cite{SRAL} showed that a Convex Optimization problem can be used to optimize the contribution weights of a set of fused image processing modalities, resulting in improved VPR performance. The optimization was computed for a separate training set for each test dataset. Multi-process fusion \cite{Hausler2019} fuses multiple modalities within a Hidden Markov Model, combining the benefits of both sequences and multiple techniques. 

More recently, evidence has been accumulating that certain combinations of modalities are more effective than others on specific datasets \cite{Hausler2020a, Waheed2021}. However, both these works rely on either a training set, or access to the full ground-truth information for a given dataset. In a different work, rather than using traditional image data, \cite{Fischer2020} fuses different temporal windows of event camera data for improved VPR. 

\subsection{Consensus Selection via Combinatorial Optimization}
In this work we propose a method for selecting the optimal subset of techniques to fuse in a multi-process fusion system. Subset selection (combinatorial optimization) has had previous investigation in the sensor fusion literature \cite{SensorSelect2009}. In that work, the sensor selection problem was relaxed from a binary selection to a soft weighted fusion, which allowed for the use of convex optimization theorems to approximate the solution to the problem without requiring a brute-force search of all combinations of sensors.

In VPR specifically, \cite{Hausler2019a} used a Greedy algorithm to find the optimal set of feature maps to use to improve localization performance, and further work used mutual information to determine the ideal subset of feature maps to use \cite{Maltar2020}. A recent work \cite{Molloy2021} intelligently selected the optimal reference set of images to use in VPR, by using Bayesian Selective Fusion. Given a set of reference images taken at different times, their algorithm found both an optimal subset of reference times and also weighted the different sets to optimize the localization performance. 

\cite{Lowry2016} used change removal to detect and remove common conditions across an environment (since common conditions are not useful at identifying unique locations \cite{Cummins2008}). Another recent work proposes an unsupervised algorithm for automatically selecting the optimal parameters for VPR algorithms \cite{Mount2021}, although only one technique was at a time. Our work can be considered a search through a large set of image representations to find the optimal subset of representations to fuse, while removing misleading representations.

\section{Methodology}

Our framework maximizes VPR performance by automatically determining the complementarity of VPR techniques and using this knowledge to fuse the complementary set of techniques in an unsupervised manner. The framework is agnostic to the set of techniques used.

\subsection{Multi-process Fusion}
In VPR, each query image observed by the autonomous platform is compared against a database of $D$ prior images using a specified feature extraction and matching technique. This produces a $D$ dimensional similarity vector, containing a list of similarity scores between the query image and each database image. A larger score denotes a stronger VPR match.

In a multi-process fusion (MPF) algorithm \cite{Hausler2019}, a variety of different VPR techniques are used simultaneously and a fusion of similarity scores can improve the overall VPR performance of a system. Formally, given a set of techniques $\mathbf{N} = \{1, \dots , N\}$, each technique $n \in \mathbf{N}$ separately computes a similarity vector $D_{n}$ containing the similarity scores between a query image and a set of database images. Since the set of techniques is arbitrary, the distribution of scores within each technique will not be consistent with the distribution in other techniques. Therefore, we use a normalization process prior to fusing the set of $\mathbf{N}$ techniques, so that each similarity vector has a minimum value of zero and a maximum value of one:
\begin{equation}
    \hat{D}_{n}(i) = \frac{D_{n}(i) - \min(D_{n})}{\max(D_{n}) - \min(D_{n})} \quad \forall i ,\quad n \in \mathbf{N}
\end{equation}

The collection of similarity vectors $\hat{D}$ are then summed to produce a combined similarity vector:
\begin{equation}
    D_\text{MPF} = \sum_{n=1}^{N} \hat{D}_{n}
\end{equation}
The matching image $X$ is then the image with the maximum score in $D_\text{MPF}: X=\argmax D_\text{MPF}$.

While such MPF algorithms have previously been shown to outperform single technique baselines, typically only a sub-selection of techniques will be suited to a particular dataset. The drastic difference in performance between different datasets and even within different sections of the same datasets of particular methods was shown in~\cite{Zaffar2021,Schubert2020}. Therefore, the inclusion of poorly performing techniques (which is the case for standard MPF) can potentially offset any advantages of using multiple methods simultaneously. Instead, in the following we describe how the best set of techniques can be found \emph{dynamically, online and without requiring training}.

\subsection{Finding Optimal Pairs of Techniques by Estimating Perceptual Aliasing}

In this subsection we begin by describing how a pair of techniques can be selected, from the set of all possible combinations of pairs out of a collection of individual techniques. Our selection criteria is based on an estimate of the perceptual aliasing of a similarity vector. In the subsequent section we describe how our approach can be scaled to larger subsets of techniques.

The intuition behind our proposed dynamic multi-process fusion is as follows. The aim of any VPR system is to find the ground-truth match in the reference database given a query image. The difficulty lies in the problem that in a sufficiently large reference database there will be images that have similar appearance that can cause mismatches -- an effect called perceptual aliasing. Our hypothesis is that there are complementary techniques in which the locations of perceptual aliasing differs, but which have correlated similarity scores around the ground-truth database image. Therefore, by estimating the perceptual aliasing of different sets of fused techniques, we can find the set of techniques that minimises this aliasing.

As established in prior literature \cite{MM2012, Hausler2019}, this estimate can be computed by finding the ratio between the similarity score at the best matching database image (the best guess for the ground-truth database image) and the score at the next highest similarity score in the database (the perceptually aliased image) -- a larger ratio denotes reduced perceptual aliasing. Unlike prior literature, which calculates this ratio score for individual techniques, we calculate this ratio for similarity vectors that are the addition of individual technique vectors.

We thus estimate the perceptual aliasing score for a pair $(k, l)$ of techniques by finding the ratio between the maximum value in the summed vector and the next largest value outside a window around the maximum value. As per:
\begin{equation}
    \hat{D}_{k+l} = \hat{D}_{k} + \hat{D}_{l}
    \label{eq3}
\end{equation}
\vspace*{-0.3cm}%
\begin{equation}
    S_{k,l} = \frac{\max(\hat{D}_{k+l})}{\max (\tilde{{D}}_{k+l})} \quad k \in \mathbf{N},\ l \in \mathbf{N},\ k\neq l
    \label{eq4}
\end{equation}
where $\tilde{D}$ denotes the subset of $\hat{D}$ that excludes all values $\pm R_\text{window}$ around $\argmax(\hat{D})$. The chosen set of techniques $(k^*, l^*)$ is the pair with the highest score $S_{k^*,l^*}$.

\subsection{Expanding to Triplets and Larger Subsets}

This scoring algorithm can be scaled up to analyze any larger combination of techniques. To do so, we find the maximum score across any combination and number of techniques allowable by the total number of techniques, all the way up to a sum of all techniques. Formally, the power set $P(\mathbf{N})$ includes all possible subsets of the set $\mathbf{N}$. We exclude the empty subset and all subsets containing only a single technique. The number of subsets is thus given by $M = 2^{N} - N - 1$.

We then re-write Eq.~(\ref{eq3}) in the general case:
\begin{equation}
    \hat{D}_{\mathbf{M}} = \sum_{m \in \mathbf{M}} \hat{D}_{m} \quad \mathbf{M} \in P(\mathbf{N})
\end{equation}
For each $\hat{D}_{\mathbf{M}}$, we then calculate the score $S_{\mathbf{M}}$ following the principle outlined in Eq.~(\ref{eq4}) and find the set of techniques $\mathbf{M}^*$ with the maximum score:
\begin{equation}
\mathbf{M}^* = \argmax (S_{\mathbf{M}}) \quad \forall \ \mathbf{M} \in P(\mathbf{N}).
\end{equation}

While this extensive approach that compares all possible variations is guaranteed to find the set of techniques that maximizes the ratio score, we note that such a search will become intractable with large values of $N$ ($N >> 10$). For such cases, the problem can be approximated by considering the technique selection problem as a convex relaxation selection process, which can be solved using interior point methods \cite{SensorSelect2009}. We leave such a solution to future work.

\subsection{Complementary-Aware Online Calibration}

As our complementary scoring technique does not require access to ground truth data, the calibration can be performed at deployment time and be continuously updated to reflect changes to the observed environment during deployment. We re-run the calibration procedure every $F$ frames, where a larger value of $F$ reduces the computational requirements. For query frames that are not used for calibration, rather than extracting features for all $N$ techniques, instead we only extract and match features for the best subset techniques decided by the calibration.

A place recognition decision is determined based on the maximum score in the summed similarity vector, for the sum of the set of complementary techniques as decided by the online calibration. The sum of techniques is weighted by a set of technique weighting scores $w_m$:
\begin{equation}
    \overline{D} = \sum_{m \in \mathbf{M}^*} w_m \hat{D}_m,
\end{equation}
where $\mathbf{M}^*$ is the set of techniques chosen by the calibration procedure. The values of $w_m$ are determined using the same ratio calculated as shown in Eq.~(\ref{eq4}), except substituting $D_{k+l}$ for the single techniques $\hat{D}_m$. This weights techniques based on their individual confidence in their place match estimate, on a frame by frame basis.

We finally normalize this summed similarity vector to have a mean of $1$ and a standard deviation of $0$, following \cite{Hausler2020}. The best matching database index is found by:
\begin{equation}
    \overline{X} = \argmax \overline{D}
\end{equation}

\section{Experimental Method}

We demonstrate our proposed dynamic Multi-process Fusion on a collection of diverse and comprehensive benchmark datasets, as provided by the recent VPR-Bench work \cite{Zaffar2021}. These datasets are: Gardens Point Walking \cite{SN2015}, SPEDTest \cite{CZ2017}, Nordland \cite{Sunderhauf2013}, Living Room \cite{Mount2016}, Synthia \cite{Ros2016}, 17Places \cite{Sahdev2016}, Cross-Seasons \cite{Larsson2019}, Corridor \cite{Milford2013}, Tokyo 24/7 \cite{Torii2018}, ESSEX3IN1 \cite{Zaffar2018}, Pittsburgh \cite{Torii2013}, and INRIA Holidays \cite{Jegou2008}. We use the same dataset configurations and ground truth tolerances as used in VPR Bench. These datasets are highly diverse, covering a wide range of indoor and outdoor environments with both traversal and non-traversal (images in the dataset are independent to each other) datasets. A variety of illumination, viewpoint, seasonal and structural changes occur between the reference and query sets of these datasets.

For all experimental results, we use a range of different VPR techniques including both hand-crafted and deep-learnt algorithms. The techniques we use are: NetVLAD \cite{Arandjelovic2018}, RegionVLAD \cite{Khaliq2020}, CoHOG \cite{Zaffar2020}, HOG \cite{DN2005}, AlexNet, AMOSNet \cite{CZ2017}, HybridNet \cite{CZ2017}, CALC \cite{Merrill}, AP-GeM \cite{Revaud} and DenseVLAD \cite{Torii2018}. These are also the same techniques used in VPR-Bench. Dynamic MPF selects the optimal subset of techniques out of this list of techniques.

We compare against a range of benchmark techniques, including different multi-process fusion algorithms and oracle hindsight-aware solutions. Our fair benchmarks are: a randomly selected pair of techniques, full Multi-process Fusion, and Hierarchical Multi-process Fusion \cite{Hausler2020a}. Our random pair baseline selects two random techniques to fuse per query image. Full Multi-process Fusion, a simplified version of \cite{Hausler2019} without sequences, is a fusion of all techniques simultaneously, without any technique selection or weighting. Finally, Hierarchical Multi-process Fusion is a re-implementation of \cite{Hausler2020a}, except using the techniques within VPR Bench. We use a three tier hierarchy with $3$, $3$ and $4$ techniques per tier respectively and the ordering of techniques is randomly assigned.

Our ``unfair'' benchmarks are configured by selecting the techniques that perform best on the experimental datasets using ground-truth information -- hereafter defined as the oracle. Our first oracle benchmark is the best performing individual technique for each dataset. The second benchmark is the best static pair of techniques, based on the average performance of the pair across all datasets -- NetVLAD and DenseVLAD were chosen. The third oracle benchmark is the best static quadruplet of techniques, which we found to be the set NetVLAD, DenseVLAD, RegionVLAD, and Ap-GeM.

We use the Recall@K metric to evaluate the VPR performance, as commonly done in the literature \cite{Arandjelovic2018, Hausler2020a}.

Hyperparameters were configured as follows. $R_\text{window}$ was set heuristically on the basis of the acceptable ground truth tolerance of each dataset. For all results, unless otherwise specified, we use a frame separation value $F$ of one.

\section{Results}
\subsection{Comparison to State-of-the-art}

In Table \ref{tab:results}, we show the recall@1 of dynamic MPF versus the fair baselines, for all datasets. While the localization performance varies across the datasets, on average, dynamic MPF outperforms all baselines with a Recall@1 of 79\% versus a Recall@1 of 74\% for Full MPF. In particular, Dynamic MPF outperforms Full MPF by large margins on the datasets Pittsburgh, ESSEX3IN1 and Tokyo247. This is because some individual techniques perform poorly on these datasets, thus reducing the overall effectiveness of the recall performance of the Full MPF baseline. In fact, on the Pittsburgh dataset five techniques out of ten have a Recall@1 less than 5\%. Dynamic MPF is selecting the best subset of techniques for each query image, which prevents poorly performing individual techniques from negatively impacting the VPR system.

Another advantage of dynamic MPF is the consistency in the recall across the datasets -- compared against all the benchmarks without the hindsight advantage, dynamic MPF is either the best performing system or a close second best performing system. Across the datasets, dynamic MPF has the best performance on the Tokyo247 dataset, with a Recall@1 of 78\% versus the benchmarks with 34\%, 67\% and 41\% for Random pair, Full MPF and Hier MPF respectively. We hypothesize that the combination of severe appearance change (day to night) and severe viewpoint change requires a VPR solution that can suit the given query image conditions. In the Tokyo247 dataset the viewpoint and appearance change is not consistent across the dataset, therefore a single, static, technique is unable to reliably perform place recognition. 

We observe that Dynamic MPF is less effective on more consistent datasets, where the appearance of the environment does not change much over the query set. This is especially the case for the Nordland and Corridor datasets, with either a static season or the same indoor building respectively. As shown in Table \ref{tab:results}, on these datasets dynamic MPF does not outperform Full MPF (however is very close in performance with a drop of just 1 and 2\% Recall@1 respectively).

\newcount\columncount
\columncount = 5

\begin{table}[!t]
  \scriptsize
  \renewcommand{\arraystretch}{1.1}
  \centering
  \caption{Comparison with the State-of-the-art:\\Recall @ 1 for all Datasets}%
  \resizebox{0.99\columnwidth}{!}{\begin{tabular}{c|cccc}
{Dataset} & {\textbf{Rand Pair}} & {\textbf{Full MPF}} & {\textbf{Hier MPF}} & {\textbf{Dyn MPF (Ours)}}\\
\cline{1-\columncount}
\cline{1-\columncount}
Gardens Point & 0.50 & 0.64 & 0.56 & \textbf{0.68}\\
SPEDTest & 0.69 & \textbf{0.85} & 0.83 & \textbf{0.85}\\
Nordland & 0.13 & \textbf{0.27} & 0.25 & 0.26\\
Living Room & 0.87 & 0.87 & 0.71 & \textbf{0.96}\\
Synthia & 0.88 & 0.91 & \textbf{0.92} & 0.91\\
17Places & 0.41 & 0.43 & 0.41 & \textbf{0.44}\\
Cross-Seasons & 0.93 & 0.98 & 0.97 & \textbf{1.00}\\
Corridor & 0.72 & 0.91 & \textbf{0.92} & 0.89\\
Tokyo247 & 0.34 & 0.67 & 0.41 & \textbf{0.78}\\
ESSEX3IN1 & 0.56 & 0.68 & 0.39 & \textbf{0.85}\\
Pittsburgh & 0.55 & 0.82 & 0.15 & \textbf{0.93}\\
INRIA Holidays & 0.78 & 0.88 & 0.81 & \textbf{0.92}\\
\cline{1-\columncount}
\textbf{Mean Recall} & 0.61 & 0.74 & 0.61 & \textbf{0.79}\\
\end{tabular}%
}
  \label{tab:results}%
  \vspace*{-0.2cm}
\end{table}%

\newcount\columncount
\columncount = 5

\begin{table}[ht]
  \scriptsize
  \renewcommand{\arraystretch}{1.1}
  \centering
  \caption{Recall @ 1 for all Datasets With Oracle Baselines That Have an Unfair Performance Advantage}%
  \resizebox{0.99\columnwidth}{!}{\begin{tabular}{c|cccc}
  {Dataset} & {\textbf{Best Indiv}} & {\textbf{NV+DV}} & {\textbf{NV+DV+RV+AP}} & {\textbf{Dyn MPF (Ours)}}\\
\cline{1-\columncount}
\cline{1-\columncount}
Gardens Point & 0.56 & 0.63 & \textbf{0.69} & 0.68\\
SPEDTest & 0.79 & 0.79 & 0.82 & \textbf{0.85}\\
Nordland & \textbf{0.29} & 0.13 & 0.14 & 0.26\\
Living Room & 0.93 & \textbf{0.96} & \textbf{0.96} & \textbf{0.96}\\
Synthia & \textbf{0.97} & 0.91 & 0.91 & 0.91\\
17Places & 0.44 & 0.44 & \textbf{0.45} & 0.44\\
Cross-Seasons & 0.99 & 0.99 & \textbf{1.00} & \textbf{1.00}\\
Corridor & 0.90 & 0.82 & 0.87 & 0.89\\
Tokyo247 & 0.60 & 0.74 & 0.76 & \textbf{0.78}\\
ESSEX3IN1 & \textbf{0.90} & \textbf{0.90} & 0.87 & 0.85\\
Pittsburgh & \textbf{0.98} & 0.96 & 0.94 & 0.93\\
INRIA Holidays & 0.92 & 0.89 & \textbf{0.94} & 0.92\\
\cline{1-\columncount}
\textbf{Mean Recall} & 0.77 & 0.76 & 0.78 & \textbf{0.79}\\
\end{tabular}%
}
  \label{tab:oracles}%
  \vspace*{-0.2cm}
\end{table}%

\subsection{Comparison to Oracles}

In Table \ref{tab:oracles}, we compare our approach with three oracle benchmarks, all utilizing prior knowledge of the datasets to make decisions on the best techniques to use. These benchmarks are the best performing individual technique, the fusion of NetVLAD and DenseVLAD (NV+DV), and the fusion of the four techniques NetVLAD, DenseVLAD, RegionVLAD and Ap-GeM (NV+DV+RV+AP). In a real deployment situation, such prior knowledge would not be available. 

Even though these baselines have this advantage, dynamic MPF still outperforms all of them, with a mean recall of 79\% versus the second highest benchmark NV+DV+RV+AP with a recall of 78\%. The main cause of reduced performance for the pair and quadruplet of techniques is the poor performance on the Nordland dataset; there dynamic MPF continues to provide high performance by utilizing other techniques that function well on this dataset. The best individual technique (again as determined on hindsight via an oracle) is a strong competitor (that would not be available on deployment), and yet the ensemble of multiple complementary techniques significantly outperforms this baseline on a number of datasets, such as Gardens Point, Tokyo247 and SPEDTest.

\subsection{Frame Separation Ablation}

To evaluate the effect of varying the calibration frequency, we ran multiple experiments with different frame separation \mbox{values $F$}. Calibrating the algorithm every $F$ frames is only sensible on sequential datasets, that is, where each query image follows the previous image. The VPR-Bench datasets that follow this format are Gardens Point Walking, Nordland, Synthia, and Corridor.%

\begin{figure}[t!]
    \centering
    \includegraphics[width=0.99\columnwidth]{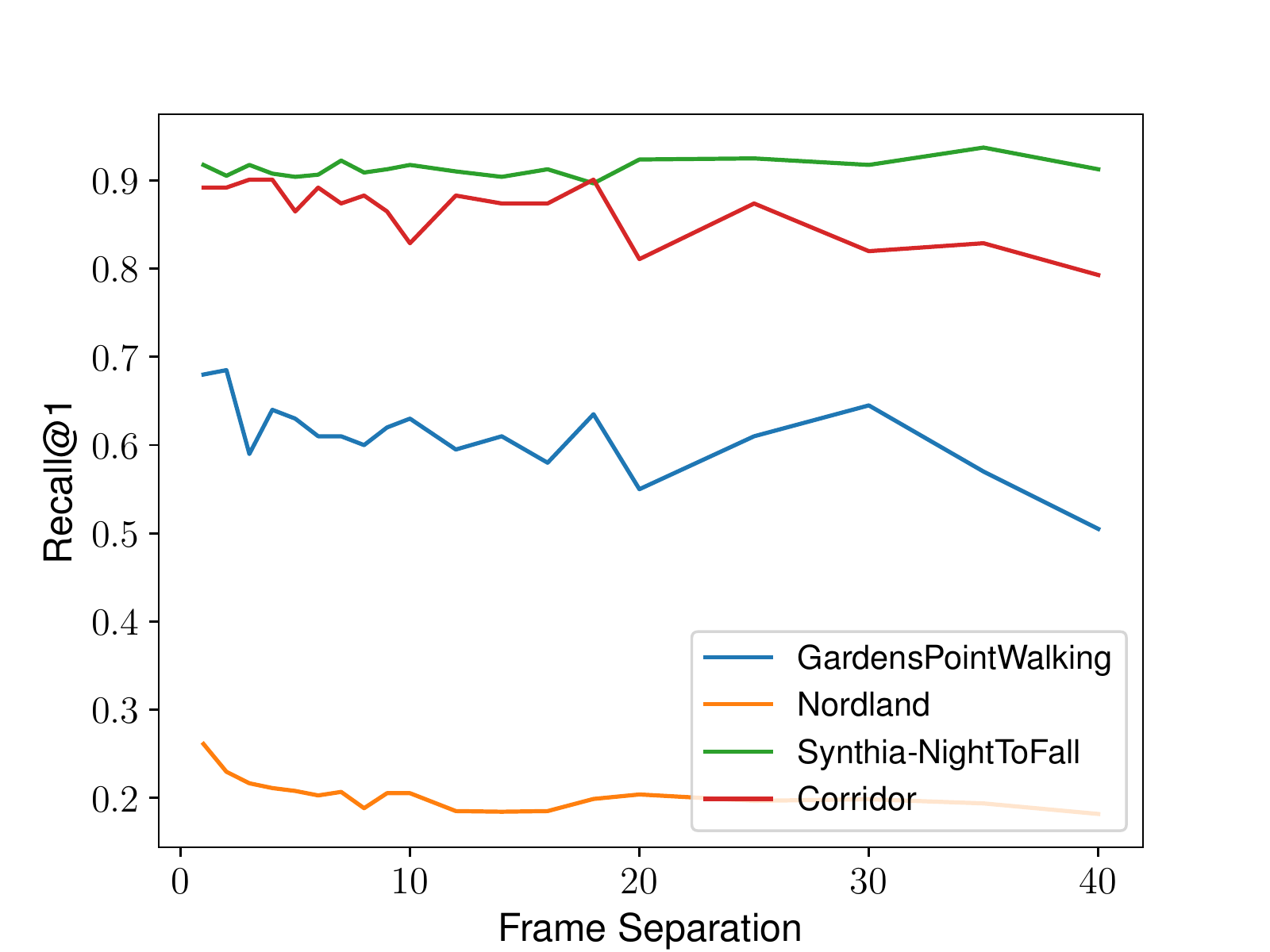}
    \caption{\textbf{Change in recall as the calibration frequency is reduced.} On the $x$-axis we show how often the calibration procedure is run, from every single frame to every $50$ frames. The resulting R@1 is shown on the $y$-axis. Performance gradually degrades as the frequency is reduced, with the reduction in performance varying by dataset.
    }
    \label{fig:framesep}
\end{figure}

Figure \ref{fig:framesep} shows the effect of varying the value of $F$. One can observe a gradual reduction in localization performance as the frame separation is increased, with different rates of reduction depending on the dataset. The Synthia dataset is an exception, which maintains recall performance even at large frame separation values. We hypothesize that the synthetic nature of this dataset means that a consistent subset of techniques is generally applicable for all the query images. For all datasets, it can be observed that even at large values of $F$, the recall only reduces by a moderate amount and the algorithm can still identify an approximately applicable subset of techniques for the environment.

\subsection{Qualitative Results}

\subsubsection{Difference Score Example}

\begin{figure}[t]
    \centering
    \includegraphics[width=\linewidth]{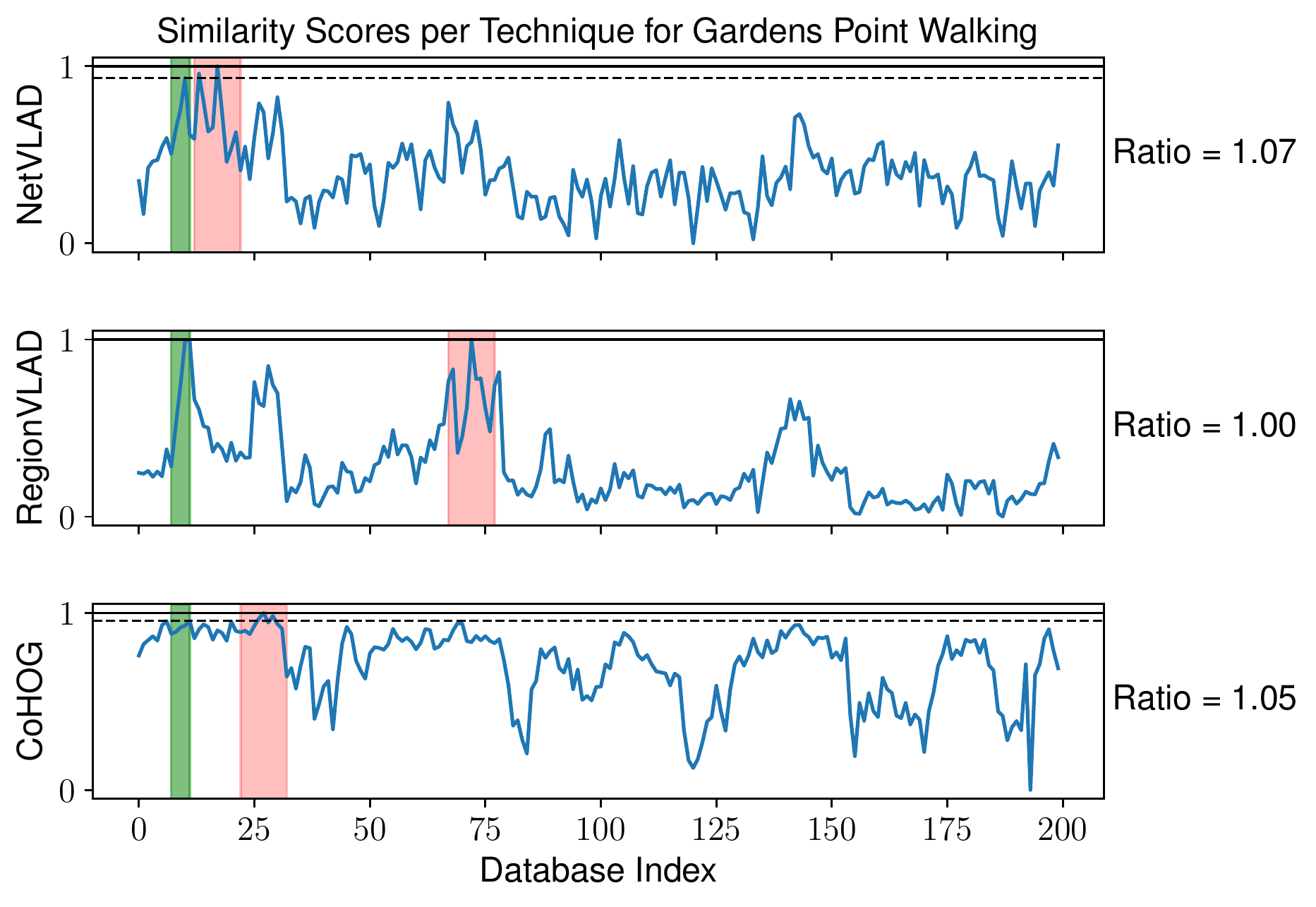}
    \caption{\textbf{Similarity scores example.} This plot shows the similarity score values for one of the query images in the Gardens Point Walking dataset (a score of $1$ denotes the database image that is most similar to the current query image). The light red bar indicates the $R_\text{window}$ area around the image that the technique has determined to be the best matching database image. The green bar indicates the ground truth matching image and the surrounding region being the images that are also within the ground truth tolerance. The ratio is calculated by dividing the score at the red bar with the next highest score outside $R_\text{window}$, denoted by the dashed line. A higher ratio is better, as this is indicative of reduced perceptual aliasing.
    }
    \label{fig:indiv_diffs}
\end{figure}

\begin{figure}[t]
    \centering  %
    \includegraphics[width=\linewidth]{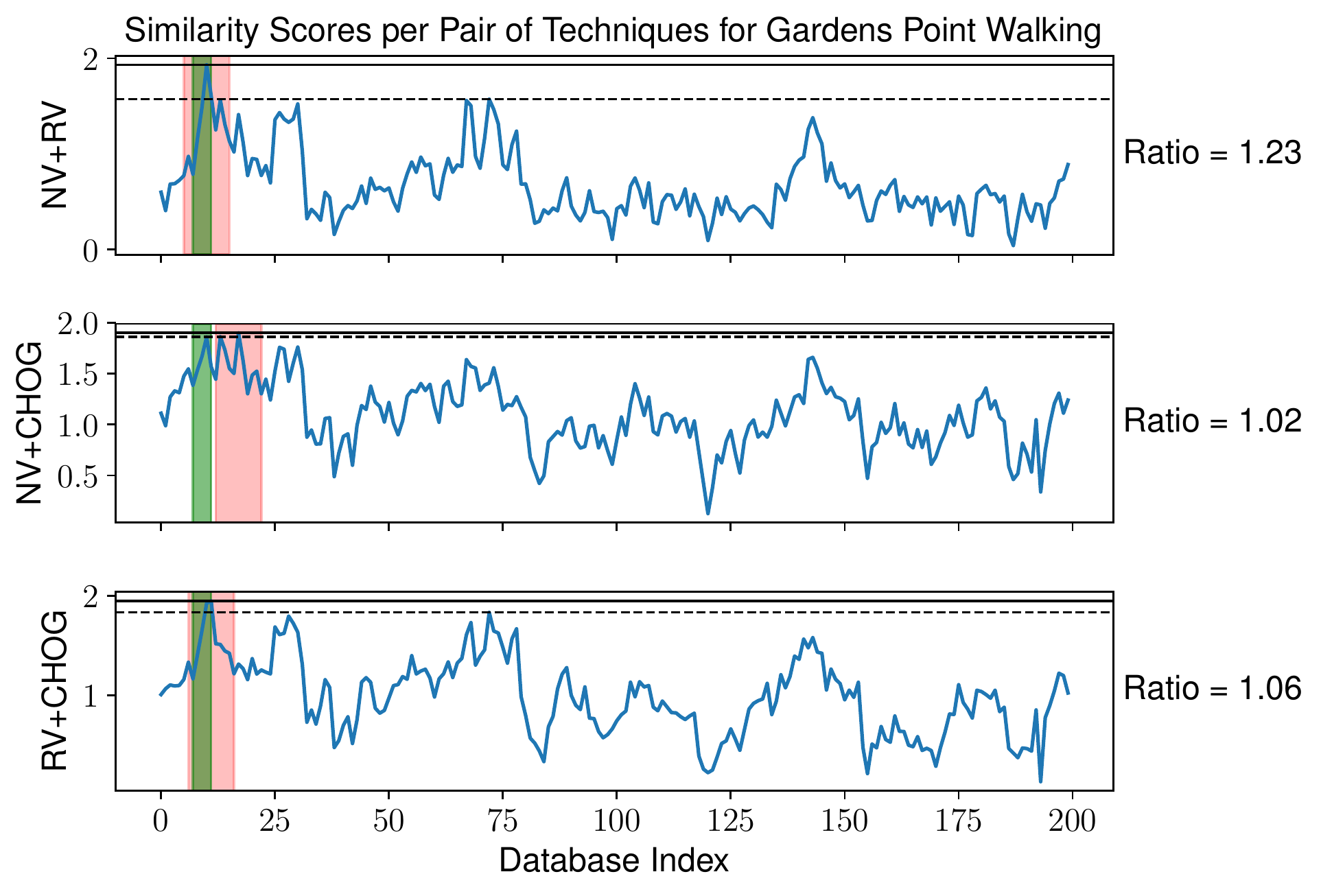}
    \caption{\textbf{Similarity score pairings.} This figure shows the similarity score for summed pairs of techniques, for all possible pairings of the individual techniques shown in Figure \ref{fig:indiv_diffs}. It can be observed that the pair with the highest ratio (NetVLAD + DenseVLAD) has correctly identified the ground truth matching database image, even though the techniques in isolation were unable to identify the correct matching image.
    }
    \label{fig:pair_diffs}
\end{figure}

In Figures \ref{fig:indiv_diffs} and \ref{fig:pair_diffs}, we show an example difference score distribution for one of the query images in the Gardens Point Walking dataset and we show how the perceptual aliasing score ratio test is calculated. In Figure \ref{fig:indiv_diffs}, none of the three individual techniques are able to identify the ground truth matching image. Severe perceptual aliasing is present, as identified by the low ratios. In Figure \ref{fig:pair_diffs}, we show all possible combinations of pairs of these techniques. Combining multiple techniques enables the correct VPR match to be found even though none of the techniques in isolation were able to identify the correct match. In this example, the pair of techniques with the highest ratio (i.e.~NV+RV) correctly identifies the database image for the current query.

\subsubsection{Techniques Selected per Query Example and per Dataset}

\begin{figure*}[p!]
    \centering
    \includegraphics[width=.74\linewidth]{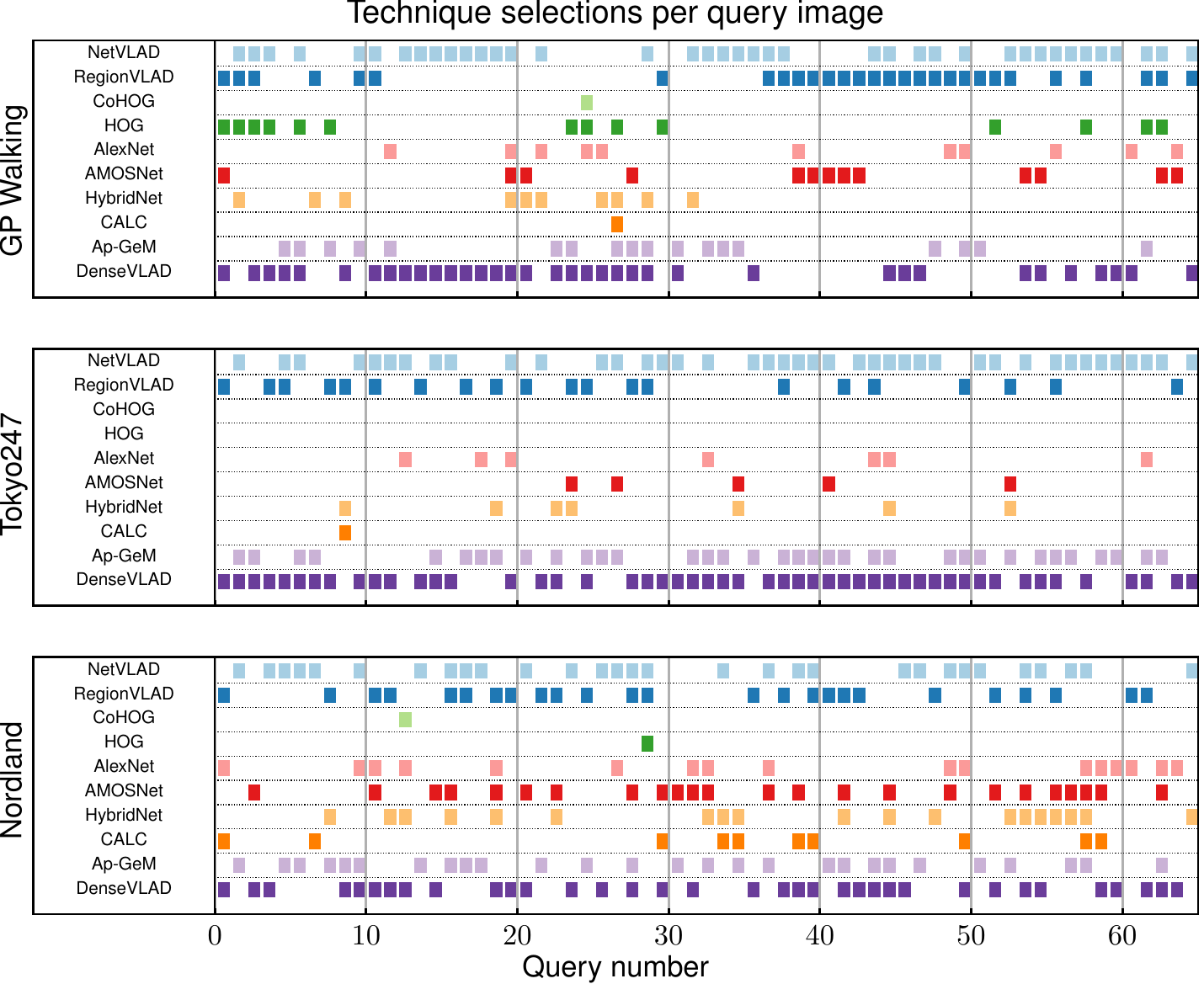}
    \caption{\textbf{Technique selections per query.} Here we show the techniques chosen for each query image using our dynamic MPF algorithm. The selected techniques varies both by dataset but also over time within datasets too. Our approach can select any number of techniques, from as little as a pair all the way up to all ten at once.
    }
    \label{fig:frameselects}
\end{figure*}

In Figure \ref{fig:frameselects}, we show the techniques selected for the first 65 query images for the datasets Gardens Point Walking, Tokyo247 and Nordland. These results indicate some interesting properties. First, the selected choice of techniques can vary significantly even between adjacent queries. This is particularly noticeable on the Nordland dataset, although we note that the real-world distance between query images is rather large on this particular dataset (approximately 200 meters). On the Gardens Point Walking dataset, we observe a slower variation in the choice of selected techniques, since adjacent queries are separated by a real-world distance of approximately just two meters.

\begin{figure*}[p!]
    \centering
    \includegraphics[width=.86\linewidth,trim=0cm 1.0cm 0cm 0cm,clip]{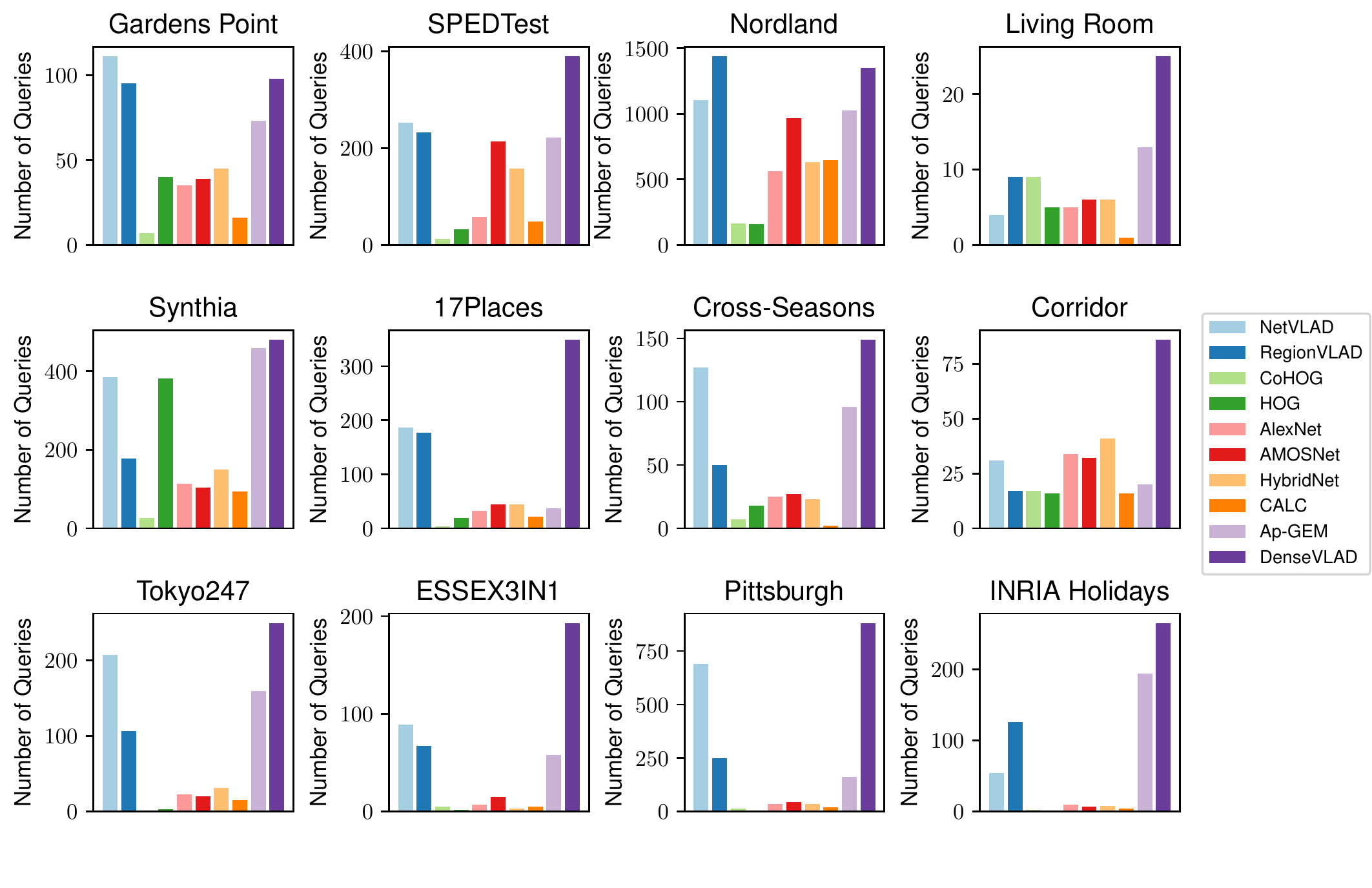}
    \caption{\textbf{Technique selection counts per dataset.} In this figure we show how often each technique is selected across each entire dataset, using our algorithm. From left to right, the techniques are: NetVLAD, RegionVLAD, CoHOG, HOG, AlexNet, AMOSNet, HybridNet, CALC, Ap-GeM and DenseVLAD. It is worth observing that the two most commonly selected techniques (across all datasets) are DenseVLAD and NetVLAD, which also corresponds to the two individual techniques with the highest average recall across the datasets.
    }
    \label{fig:techselectcounts}
\end{figure*}

Second, we note that individual datasets have ``preferred'' techniques which are more consistently selected. For example, DenseVLAD is highly preferred on the Tokyo247 dataset and AMOSNet is regularly chosen in the Nordland dataset. This is further detailed in Figure \ref{fig:techselectcounts}, which shows how often each technique is selected across one dataset. These preferred techniques indicate that these techniques are suited to the general conditions within a specific environment, such as the presence (Tokyo247) or lack of viewpoint variations (Nordland). Another example is the Pittsburgh dataset, where NetVLAD and DenseVLAD are almost always selected; this is explainable considering that NetVLAD and DenseVLAD are both somewhat viewpoint invariant, and the Pittsburgh dataset contains significant viewpoint variaions but minimal appearance change. 

\subsubsection{Match Distributions over Perceptual Aliasing Score}

\begin{figure}[ht]
    \centering
    \includegraphics[width=0.95\columnwidth,trim=1cm 0.5cm 1cm 0.5cm,clip]{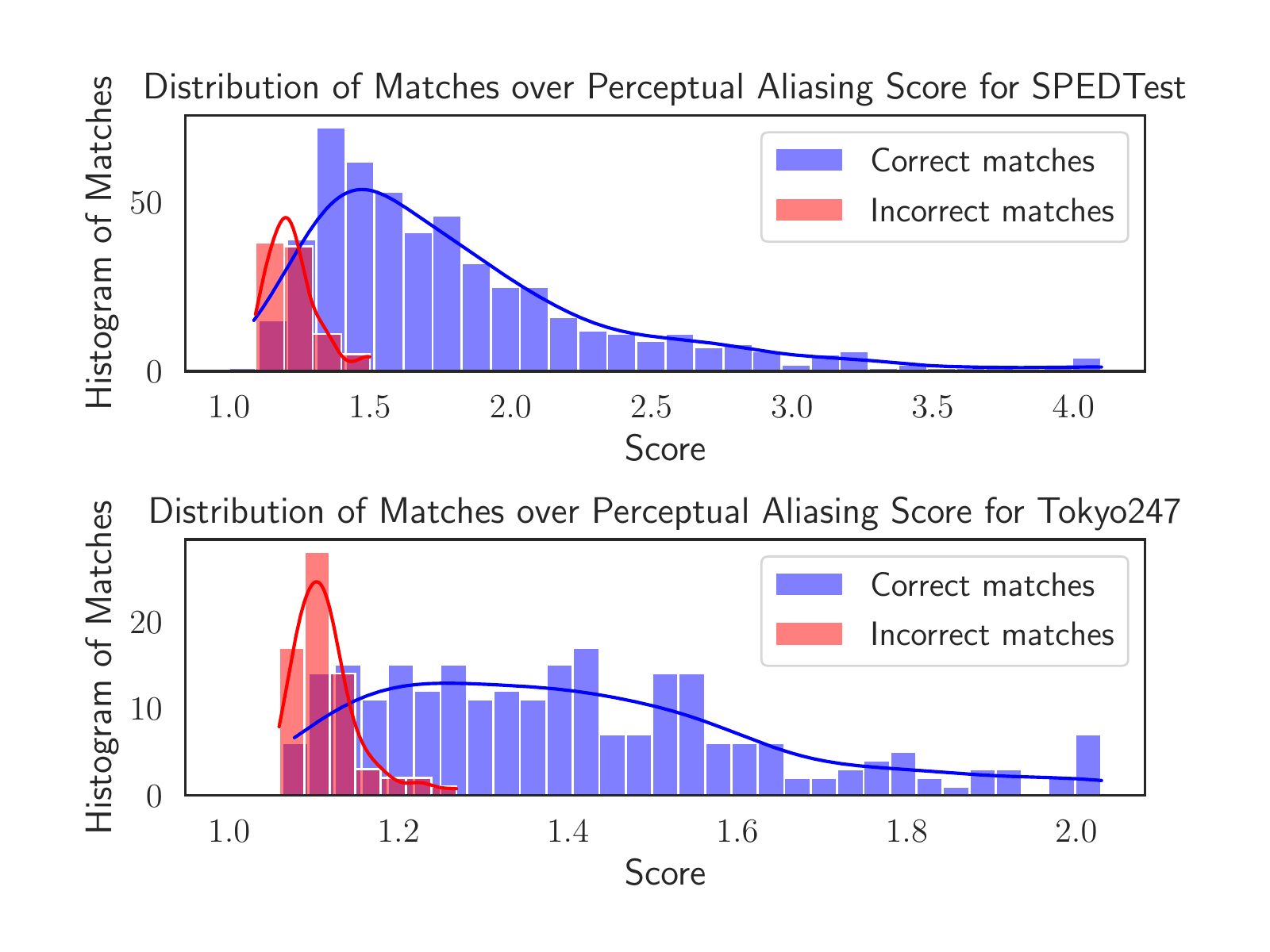}
    \caption{\textbf{Correlation between perceptual aliasing score and recall.} To demonstrate the effectiveness of our approach, this figure shows the perceptual aliasing score generated by our algorithm against a histogram of match success and failure cases. Our approach selects the subset of techniques that maximizes this score, and the recall is determined from a database search using these techniques. The figure demonstrates a clear separation between the success and failure distributions.
    }
    \label{fig:corr}
\end{figure}

To understand the correlation between our perceptual aliasing score and the recall performance of fused techniques, we plot a histogram of the score of the best subset of fused techniques (the maximum ratio value), where the number of correct and incorrect place matches are counted for a given histogram bin (for all query images in a dataset). Figure \ref{fig:corr} shows this distribution for SPEDTest and Tokyo247, with a clear distinction between correct and incorrect distributions based on our metric. This figure demonstrates that indeed there is a correlation between the perceptual aliasing and the complementarity of techniques.

\section{Discussion and Conclusion}
\label{sec:discussion}
In this work we propose an unsupervised algorithm that selects a complementary subset of VPR techniques that are best suited in the current navigation environment. We found that an estimate of perceptual aliasing, using the ratio between similarity scores during a database search, can be used to find a combination of techniques that minimises the perceptual aliasing. To the best of our knowledge, this is the first work that is able to select a complementary fusion of different VPR techniques in an unsupervised manner with frame by frame technique selections, without requiring ground-truth information or retrospective knowledge of the environment. 

Our work opens up a number of potential avenues of future work. First, convex relaxation methods can be used to scale our approach to any value of $N$. Second, deep learning methods could be used to learn an attentional model, which would learn to select the best subset of techniques. Such techniques have seen success in other fields of research, such as visual object tracking \cite{Choi2017}. %

This work provides a VPR algorithm that is able to adapt the set of techniques used to suit any dataset, as evidenced by the fact that Dynamic MPF has the highest mean recall, compared to both oracle and without oracle baselines. Such a solution makes headway towards resolving an existing problem with Multi-process Fusion algorithms, that is, not knowing which set of VPR techniques to use a-priori for the current deployment environment.

\clearpage

\IEEEtriggeratref{22}

\bibliographystyle{IEEEtran}
\bibliography{DynamicMPF}

\end{document}